\documentclass[10pt,twocolumn,letterpaper]{article}

\usepackage{arxiv}
\usepackage{times}
\usepackage{bbm}
\usepackage{epsfig}
\usepackage{graphicx}
\usepackage{amsmath}
\usepackage{amssymb}
\usepackage{color}
\usepackage[ruled]{algorithm2e}
\usepackage{booktabs}
\usepackage{multirow}

\usepackage[pagebackref=true,breaklinks=true,letterpaper=true,colorlinks,bookmarks=false]{hyperref}

\begin{document}

\title{Computer-aided Detection of Squamous Carcinoma of the Cervix in Whole Slide Images}

\author{
	Ye Tian$^{\#}$, 
	Li Yang$^{\#}$, 
	Wei Wang$^{\#}$, 
	Jing Zhang, 
	Qing Tang,\\
	Mili Ji, 
	Yang Yu, 
	Yu Li$^{\ast}$, 
	Hong Yang$^{\ast}$,
	Airong Qian$^{\ast}$,\\
}
\maketitle

\begin{abstract}
Goal: Squamous cell carcinoma of cervix is one of the most prevalent cancer worldwide in females. Traditionally, the most indispensable diagnosis of cervix squamous carcinoma is histopathological assessment which is achieved under microscope by pathologist. However, human evaluation of pathology slide is highly depending on the experience of pathologist, thus big inter- and intra-observer variability exists. Digital pathology, in combination with deep learning provides an opportunity to improve the objectivity and efficiency of histopathologic slide analysis. Methods: In this study, we obtained 800 haematoxylin and eosin stained slides from 300 patients suffered from cervix squamous carcinoma. Based on information from morphological heterogeneity in the tumor and its adjacent area, we established deep learning models using popular convolution neural network architectures (inception-v3, InceptionResnet-v2 and Resnet50). Then random forest was introduced to feature extractions and slide-based classification. Results: The overall performance of our proposed models on slide-based tumor discrimination were outstanding with an AUC scores > 0.94. While, location identifications of lesions in whole slide images were mediocre (FROC scores > 0.52) duo to the extreme complexity of tumor tissues. Conclusion: For the first time, our analysis workflow highlighted a quantitative visual-based slide analysis of cervix squamous carcinoma. Significance: This study demonstrates a pathway to assist pathologist and accelerate the diagnosis of patients by utilizing new computational approaches. 
\end{abstract}

\let\thefootnote\relax\footnotetext{
	\par This work is supported by the National Natural Science Foundation of China (No. 81872129) and Tianjin Nature Science Foundation (No. 16JCQNJC13100). \textbf{Asterisks indicate corresponding authors. Pounds indicate co-first authors.}
	\par Ye Tian, Li Yang, Wei Wang, Qing Tang, Mili Ji, Yu Li (e-mail: liyu@nwpu.edu.cn) and Airong Qian (e-mail: qianair@nwpu.edu.cn) are with Lab for Bone Metabolism, Key Lab for Space Biosciences and Biotechnology, School of Life Sciences, Northwestern Polytechnical University, Xi'an, Shaanxi 710072, China; Research Center for Special Medicine and Health Systems Engineering, School of Life Sciences, Northwestern Polytechnical University, Xi'an,Shaanxi 710072, China and NPU-UAB Joint Laboratory for Bone Metabolism, School of Life Sciences, Northwestern Polytechnical University, Xi'an, Shaanxi 710072, China. 
	\par Yang Yu is with Tianjin Key Laboratory on Technologies Enabling Development Clinical Therapeutics and Diagnostics (Theranostics), School of Pharmacy, Tianjin Medical University, Tianjin, China.
	\par Jing Zhang is with Department of Pathology, XiJing Hospital, Fourth Military Medical University, Xi'an, 710032, P. R. China.
	\par Hong Yang is with Department of Gynaecology and Obstetrics, XiJing Hospital, Fourth Military Medical University, Xi'an, 710032, P. R. China (e-mail: yanghong@fmmu.edu.cn).
}

\section{Introduction}\label{sec:intro}
Cervical cancer is ranked second or third main reason of cancer death in females, especially in economically developing countries \cite{mathew2009trends, jemal2011global, sarmadi2017association}. An estimated deaths from cervical cancer worldwide is 266,000 in 2012 \cite{international2012globocan}. adenocarcinoma and Squamous cell carcinoma (SCC) are most common types of cervical cancer \cite{wesola2015morphometric}. Especially, the former is the dominant pathological type, which constitutes more than 80\% of all cases of cervical malignancy \cite{wesola2015morphometric, fujiwara2014gynecologic}.

The routine diagnosis of squamous carcinoma of cervix including high-risk human papillomavirus (HPV) DNA detection, Papanicolaou (Pap) test and biopsy on the lesions of the cervix uteri \cite{sung2011cervical, poomtavorn2013cytohistologic, miller2015squamous}. Among them, histopathological examination is the most reliable way for SCC diagnosis, and some molecular markers have been proposed to assess the stage and prognosis of SCC by means of immunohistochemistry \cite{zhan2014expression, ravarino2012cintec}. In conventional practice of visual examination, histological tissue is stained with haematoxyline and eosin (H\&E) and then detected, identify and graded under microscope. Morphological interpretation of histological sections casts the foundation of SCC diagnosis and prognostication. However, it is subjective that visual scoring of SCC, and accordingly inclined to inter- and intra-observer variability. Moreover, due to the rise in cancer incidence and personalized medicine need, Identify and diagnosis SCC has become more and more difficult \cite{litjens2016deep}. The modern pathologists need analyze various slides, and extract quantitative parameters (e.g. surface areas, lengths, mitotic counts) quickly to come to a complete diagnosis. Thus a more objective, accurate, standardized and reproducible method is required to fulfill the daily clinical routine. 

With the progress of slide scanning technology, it is become possible to full digitalization of the microscopic analysis of stained tissue sections in histopathology recently. Particularly, computer histological analysis has turned into one of the most fascinating domain in medical imaging process \cite{goode2013openslide}. Computer-aided diagnosis not only can ease the burden on pathologists', but also benefit to eliminate the misdiagnosis rate and inconsistency between different observers. Convolution neural networks (CNNs) already been extensively used for numerous imaging process including histopathological slides because of its efficient structure in deep learning \cite{zhang2017recent, cai2017improving, maninis2016deep, hou2016patch}. Some investigations have proved CNNs as promising tools for the detections of whole slide images (WSIs) which are extremely large \cite{hou2016patch, liu2017detecting, vang2018deep}. Usually, small patches (e.g. 256 $ \times $ 256 pixels) were extracted from WSIs as the training data, and then a CNN has been trained with the training data above for the classification of patches. Finally, probability maps of the test WSIs were acquired according to the prediction results of patches, and cancer inspections were executed based on those probability maps. 

In this study, we explored deep CNNs to analyze WSIs of cervix squamous carcinoma which never been detected by computer due to complexity of the specimens. We compared the accuracies of different state-of-the-art CNN methods for cervix squamous carcinoma diagnosis. After fine training, all methods could discriminate tumor existence in the tested WSIs. But the locations of tumors were not predicted accurately enough indicating that more suitable algorithms need to be established for cervix squamous carcinoma detection. And the widely use of deep CNNs in WSIs will likely raise diagnostic accuracy and efficiency of cervix squamous carcinoma.

\section{Materials And Methods}
\begin{figure}
	\begin{center}
		\includegraphics[width=1.0\linewidth]{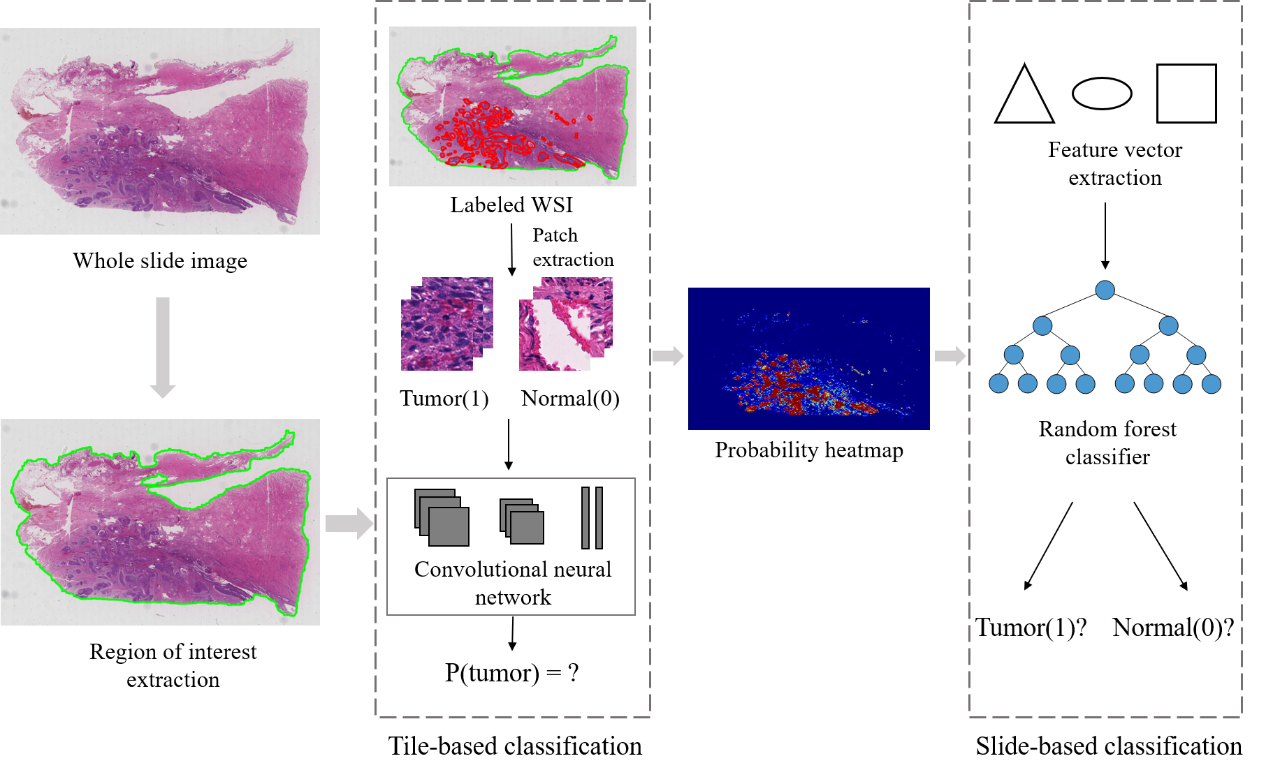}
	\end{center}
	\caption{Overall architecture of a cervical cancer prediction framework}
	\label{fig:overall_structure}
\end{figure}

Our materials and approaches are described in this section and the overall architecture of cervical cancer prediction framework is shown in Fig.~\ref{fig:overall_structure}. In detail, we first introduce the cervical cancer histopathology image acquisition and its ground truth. We then present the regions or interest detection with image processing and the deep learning algorithms for tile-based classification. Finally, the post-processing on heatmaps for slide-based classification was detailed. 

\subsection{Histopathology Image Acquisition and Ground Truth}
\begin{figure}
\begin{center}
\includegraphics[width=1.0\linewidth]{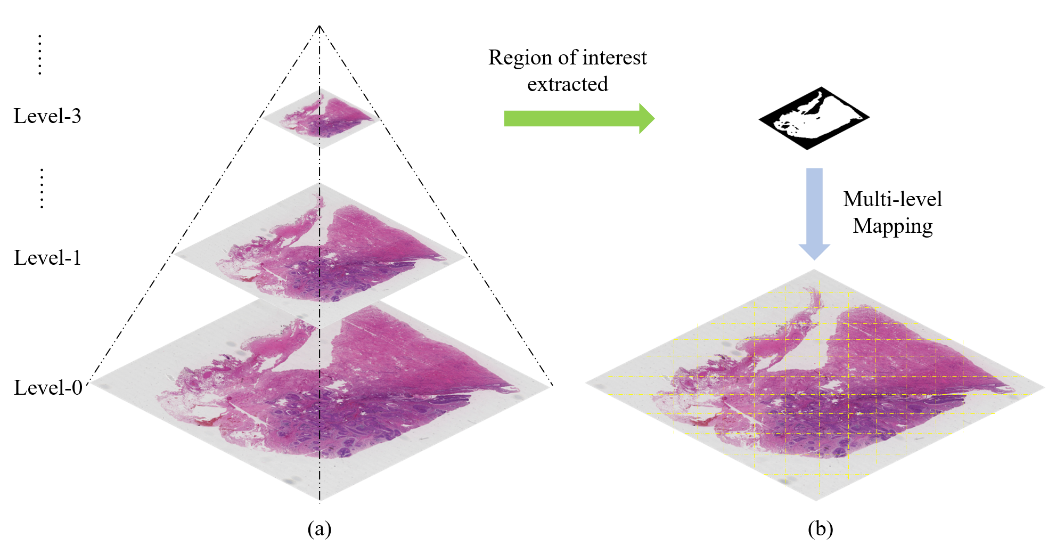}
\end{center}
   \caption{The WSI architecture and multi-layer mapping approach. (a) The multi-resolution pyramid architecture of WSI. (b)multi-layer mapping approach for regions of interest detection.}
\label{fig:pyramid_ar}
\end{figure}

A total of 800 H\&E stained slides of cervix squamous carcinoma from 300 patients were collected in Department of Pathology, Xijing hospital, Fourth Military Medical University between June 2015 and December 2017. H\&E stained pathological images were scanned to WSIs by a whole-slide scanner KF-PRO-120 (Konfoong Biotech International Co., Ltd.) which has a high-resolution. As a result, each H\&E stained section has digitized to a gigapixel image of $ \times $4 to a $ \times $40 magnification with multiresolution pyramid architecture (Fig.~\ref{fig:pyramid_ar}{\color{red}(a)}). WSIs were further annotated under the supervision of. 4 students further annotate the WSIs at the pixel level to distinguish between cancer and normal tissue and then every WSI was confirmed by 2 expert pathologists (Department of Pathology, Xijing hospital, Fourth Military Medical University).

\subsection{Regions of Interest (ROI) Detection with Image Processing}
WSIs are large GigaPixel (106$ \times $106 pixels) images. Thus, it's a super time-consuming task if the whole region of WSI is dealt with. Therefore, the first stage in our pipeline is to extract tissue region from whole slide image by removing the background (white space). ROI detection is a critical step because it helps to reduce computation time by only dealing with regions where tumor is more likely to occur. So as to extract tissue regions from the whole slide images, OTSU \cite{otsu1979threshold} algorithm was applied using a multi-layer mapping approach due to the benefit of pyramid architecture, as shown in Fig.~\ref{fig:pyramid_ar}{\color{red}(b)}. In detail, we first convert the original image from RGB color space to HSV (hue, saturation and value) color space as analyzing color values is more convenient in HSV color space. Then, a binary mask is generated by using Otsu’s threshold-based technique. Finally, tissue regions are extracted using the tissue mask which is obtained by opening and closing image morphology operations in binary mask. 

\subsection{Deep-learning CNN for Tile-based Classification}
\begin{figure*}
	\begin{center}
		\includegraphics[width=1.0\linewidth]{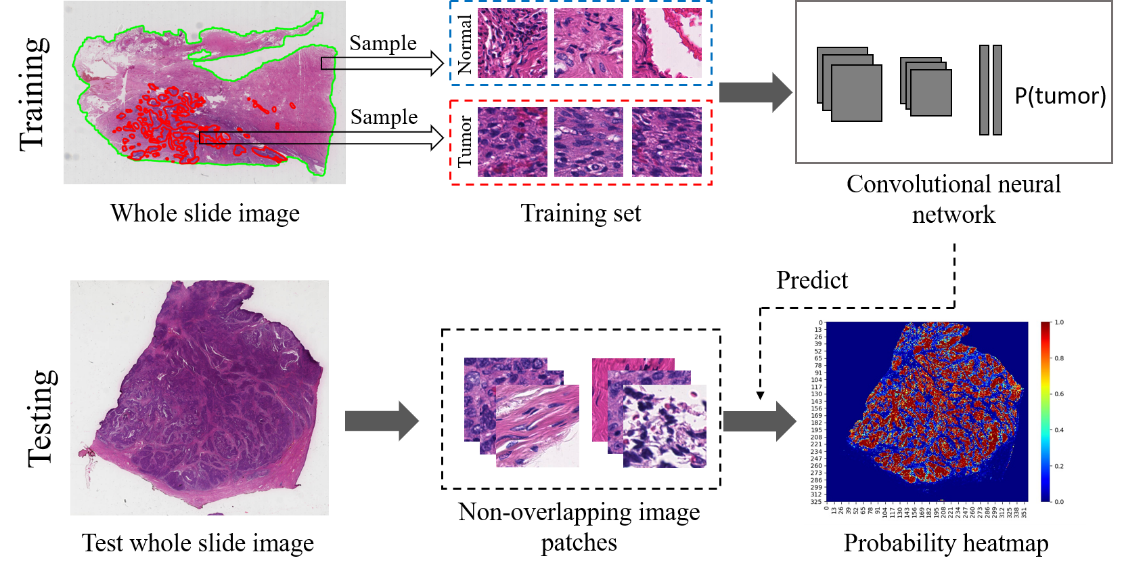}
	\end{center}
	\caption{Tile-based classification framework.}
	\label{fig:tile_framwork}
\end{figure*}

Our tiled-based classification framework as shown in Fig.~\ref{fig:tile_framwork} is implemented using Keras library on the workstation equipped with a 16 GB NVIDIA Quadro GP100. Our goal is to identify the tumor existence and location in cervical cancer WSI, then present to pathologist review. Considering the huge size of WSI and the limited memory of computer, the CNN model is trained using small patches (256$ \times $256 pixel) instead which are extracted from the slide to reduce the computing time \cite{litjens2016deep, fy201510}. Because a 256-pixel region already span many tumor and/or normal cells. Finally, we randomly extracted thousands of 256$ \times $256 pixel size patches from ROIs of cervical cancer WSI images.  

Our tumor area detection module is based on the popular CNN architectures inception-v3 \cite{szegedy2016rethinking}, InceptionResnet-v2 \cite{szegedy2017inception} and Resnet50 \cite{he2016deep} for patches classification to identify between tumor and normal patches. During training stage, 256$ \times $256 pixel patches from positive and negative regions were used as input to train classification models to distinguish the tumor patches from normal patches. We trained those three models using the same extracted patches and obtain model A (inception-v3), model B (InceptionResnet-v2) and model C (Resnet50) respectively. Data augmentation techniques have been applied to combat with the variety of H\&E by haphazardly rotating, random cropping 224$ \times $224 pixel from original 256$ \times $256 pixel patches and randomly left-right flipping \cite{RN24}. In addition, we ensembled those three trained models to optimize the performance and named it as model D. 

\subsection{Post-processing on Heatmaps for Slide-based Classification}
By using the patches classifier, every WSI is transformed to a probability heatmap in which a 256$ \times $256 pixel patch is considered as a single pixel. Post-processing operations are carried out on these heatmaps to build the classifier for slide-based classification. In short, features extracted from  heatmaps for its corresponding WSI is taken as input, and after processing, a single predicted label of the full whole slide image is outputted. Initially, 28 morphological and geometrical features, including the longest axis of the tumor region, the area ratio between tumor region and the minimum surrounding convex region, the percentage of tumor region over the whole tissue region and the average prediction values, are extracted from each heatmap \cite{liu2017detecting}. Then Random Forest classifier is constructed and applied using those features extracted above to discern the tumor whole slide images from the normal whole slide images. The whole framework of slide-based classification is shown in the slide-based classification section of Fig.~\ref{fig:overall_structure}. 

\section{Results}\label{sec:3}
\subsection{Patient Characteristics and Dataset}
A total of 800 WSIs from a cohort of 300 patients (Tab .~\ref{tab:suplement}) were collected from Department of Pathology, Xijing hospital (Fourth Military Medical University), encompassing cervix squamous carcinoma as well as adjacent benign tissue. These WSIs were further assigned into Training Set, Cross Validation Set and Test Set (Tab.~\ref{tab:the_dataset}). The models were firstly trained using Training Set, then examined in Cross Validation Set to optimized training procedure but avoiding overfitting. Finally, these models were employed to Test Set evaluation.
\begin{table}[]
	\begin{center}
		\setlength{\tabcolsep}{1.2mm}{
			\begin{tabular}{lcccc}
				\toprule
				\multirow{2}{*}{\textbf{}} & \multirow{2}{*}{\textbf{Training set}} & \textbf{Cross validation} & \multirow{2}{*}{\textbf{Test set}} & \multirow{2}{*}{\textbf{Total}}\\
				  &  & \textbf{set} &  & \\ \hline
				Tumor(1) & 204 & 74 & 69 & 347 \\
				Normal(0) & 276 & 86 & 91 & 453 \\
				Total & 480 & 160 & 160 & 800 \\
				\bottomrule& 
		\end{tabular}}
	\end{center}
	\caption{The dataset.}
	\label{tab:the_dataset}
\end{table}

\subsection{Implementation details}
For tiled-based classification, 256$ \times $256 pixel patches were extracted from whole slide images at the highest magnification ($ \times $40) level with 1.20 $ um $/pixel resolution. On average, 1000 tumor and 500 normal patches from each Tumor slide, and 500 normal patches from each Normal slide were obtained respectively. In total, 444k patches are extracted, in which 204k were tumor patches and 240k were normal patches. For training, Adam was used as optimizer, Softmax cross entropy acted as the loss function, exponential decay mechanism for learning rate management, and 32 is the batch size. 0.01 was set as the initial learning rate. In order to reduce oscillation and divergence during model training, the learning rate value was decreased by the factor of 10 when the training loss cannot reduce. By applying this setup, we trained popular CNN models (ResNet50, Inception-v3, InceptionResnet-v2) until them converge respectively. 
  
Then, random forest was introduced to features extracted from heatmaps associated with WSIs. The features we used were the same as previous work \cite{chen2016identifying, wang2016deep}. And after analyzing the importance of individual features, top 5 important features have been selected for the input of our random forest model. The top 5 important features are demonstrated in Tab.~\ref{tab:top_features}. 
\begin{table*}
	\begin{center}
		\setlength{\tabcolsep}{1.2mm}{
			\begin{tabular}{lcc}
				\toprule
				& \textbf{Feature ($ t $ is probability threshold)} \\ \hline
				1 & Mean area of tumor region ($ t = 0.9 $) \\
				2 & The longest axis in the largest tumor region ($ t = 0.5$) \\
				3 & Ratio of pixels in the region to pixels in the total bounding box ($ t = 0.5 $) \\
				4 & Eccentricity of the ellipse that has the same second moments as the region. ($ t = 0.9 $) \\
				5 & Ratio of tumor region to the tissue region ($ t = 0.9 $) \\
				\bottomrule&
		\end{tabular}}
	\end{center}
	\caption{Top 5 important features extracted from a heatmap prediction.}
	\label{tab:top_features}
\end{table*}

\subsection{Evaluation Results}
\begin{figure}
	\begin{center}
		\includegraphics[width=1.0\linewidth]{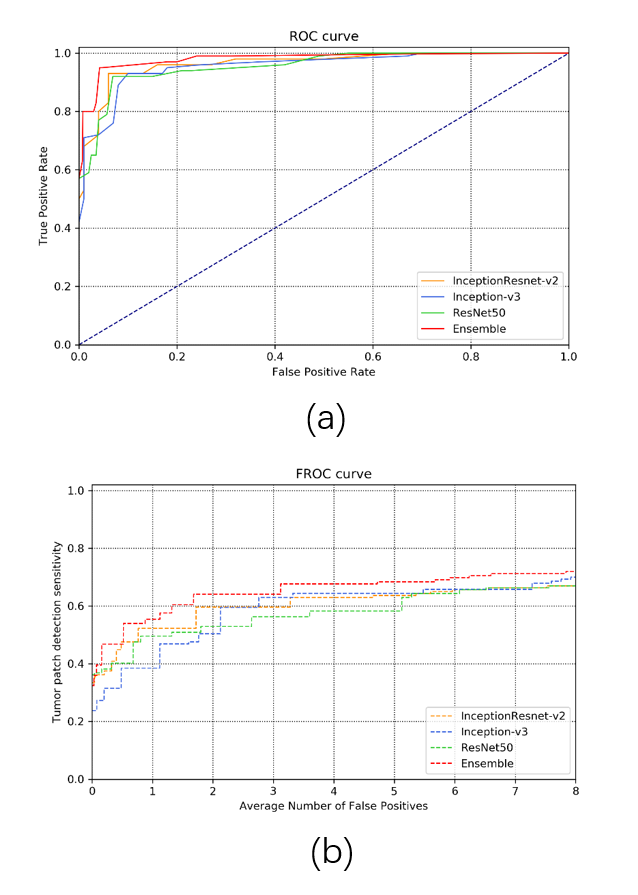}
	\end{center}
	\caption{Evaluation Conclusions. (a): FROC curves of tumor region localization of different models. (b): ROC curves of slide-based classification task of different models.}
	\label{fig:results_roc}
\end{figure}

The challenges were laid on two tasks which were WSI classification and tumor region localization. The former was to discriminate between whole slide images containing tumor regions and normal whole slide images. Receiver operating characteristic (ROC) analysis were conducted and the AUC (area under the curve) of ROC was used to evaluate the slide-level algorithms quantitatively \cite{hajian2013receiver}. We evaluated the performance of three well-known deep CNN architectures (Inception-v3, InceptionResnet-v2 and Resnet50) and our ensemble model for this classification task. As Tab .~\ref{tab:results} and Fig.~\ref{fig:results_roc}{\color{red}(a)} shown, all models exhibited good performance on patch-based classification, which indicated fine tumor recognition abilities. Our ensemble model (model D) received the highest score.
\begin{table}
	\begin{center}
		\setlength{\tabcolsep}{1.2mm}{
			\begin{tabular}{lccc}
				\toprule
				\textbf{Model} & \textbf{AUC score} & \textbf{FROC score} \\ \hline
				Inception-v3 (model A) & 0.9476 & 0.5279 \\
				InceptionResnet-v2 & \multirow{2}{*}{0.9621} & \multirow{2}{*}{0.5417} \\
				(model B) &  &  \\
				Resnet50 (model C) & 0.9593 & 0.5338 \\
				Ensemble (model D) & 0.9784 & 0.5609 \\
				\bottomrule&
		\end{tabular}}
	\end{center}
	\caption{Evaluation of various deep models. }
	\label{tab:results}
\end{table}

\begin{figure}
	\begin{center}
		\includegraphics[width=1.0\linewidth]{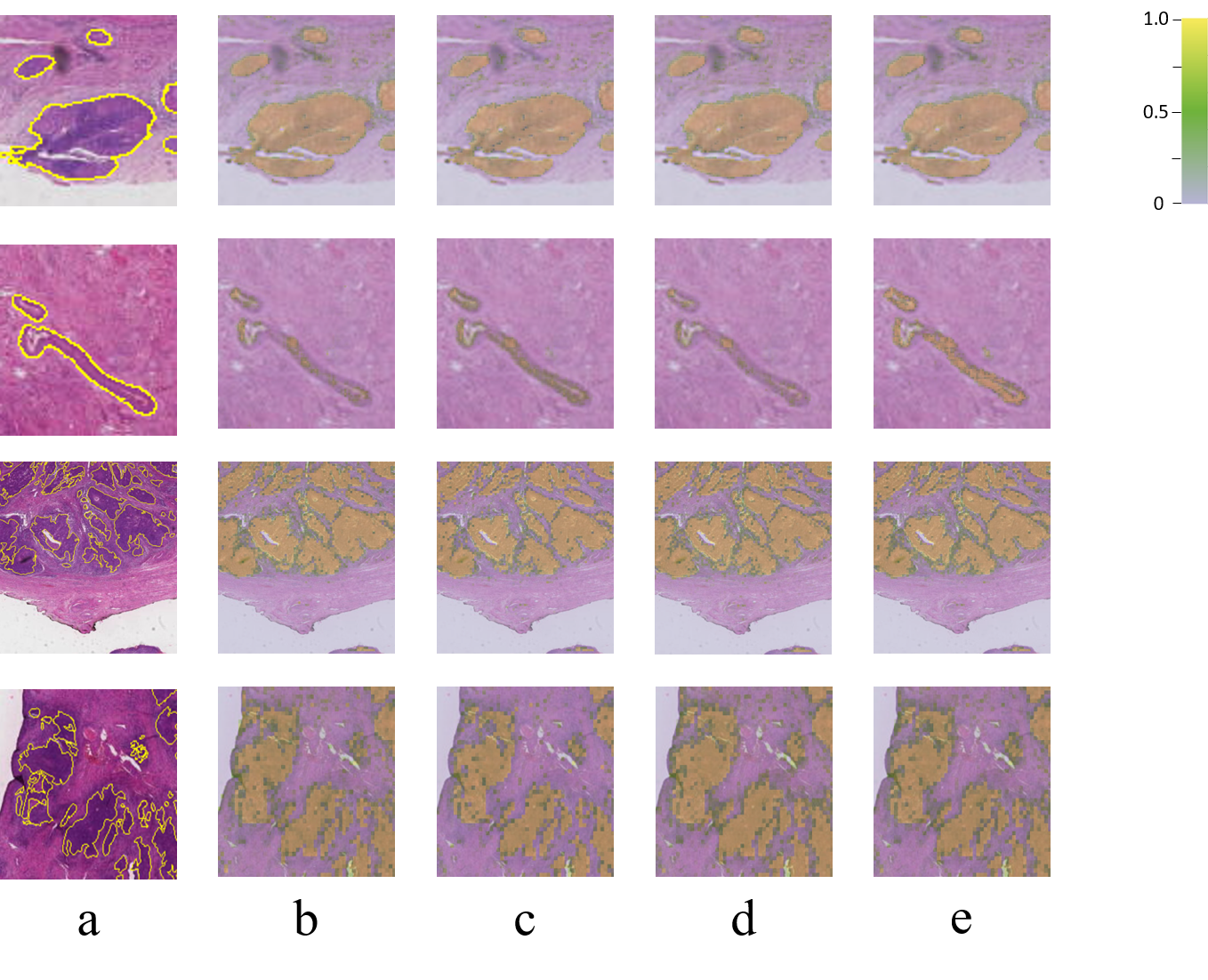}
	\end{center}
	\caption{Visualization of cervical cancer tumor region detection. (a) represents the WSIs and its ground truth. And (b), (c), (d), (e) represent the heat map generated by model A, model B, model C and model D, respectively.}
	\label{fig:visul_results}
\end{figure}

The tumor region localization, was assessed by Free Response Operatinh Characteristic (FROC) curve \cite{chakraborty1989maximum}. The FROC curve is defined as the plot of sensitivity versus the average number of false-positives per image. Our results were exhibited in Tab .~\ref{tab:results} and Fig.~\ref{fig:results_roc}{\color{red}(b)} and Qualitative results are shown in Fig.~\ref{fig:visul_results}. Among them, the model D has achieved the best performance in tumor region localization. All of the model, include model A, B, C and model D, also have exciting results in visualization of tumor region localization. The Fig.~\ref{fig:visul_results}{\color{red}(a)} show the WSI and its correspond ground truth, followed by heatmaps generated on model A, model B, model C and model D in Fig.~\ref{fig:visul_results}{\color{red}(b)} to Fig.~\ref{fig:visul_results}{\color{red}(e)}, respectively. The red region in Fig.~\ref{fig:visul_results}{\color{red}(a)} present cervix cancer region. As can be observed from Fig.~\ref{fig:visul_results}{\color{red}(b)}, Fig.~\ref{fig:visul_results}{\color{red}(c)} and Fig.~\ref{fig:visul_results}{\color{red}(d)}, the model A, model B and model C have a high false positive rate. Model D have reduced the high false positive rate which was seen in Fig.~\ref{fig:visul_results}{\color{red}(e)}. Those visualization results validate that our system makes the diagnosis decisions based on real discriminative regions. 

\section{Discussion}
The medical specialty of pathology shoulders the responsibility of disease diagnoses to guide therapy. Accurate, reproducible and standardized pathological diagnoses plays a significant role in the development of precision medicine. Owing to the limitation of microscopic images analysis, such as cognitive load, diagnostic errors and non-standardization, there has been increasing interest in developing computational methods to aid histopathological slide evaluation over the several past decades \cite{ghaznavi2013digital, gurcan2009histopathological, hipp2011computer}. In previous years, the computing methods were put forward based on prior knowledge about the target structures and shapes \cite{bentaieb2016topology, casanova2017morphoproteomic, qi2011robust}, because carcinogenesis produces characteristic morphologic changes in cancer cells \cite{muller1973nature}. And several traditional machine learning frameworks were applied in digital histopathology, e.g. Support Vector Machine for gland detection in prostate cancer \cite{tabesh2007multifeature}. While hand-crafted features which were summarized from prior knowledge of tumor morphology were too limited to represent the complex and high dimension features. In the near future, deep learning algorithms have achieved significant success in medical image recognition tasks, especially for WSI analysis of tumor \cite{wang2016deep, kermany2018identifying, li2018cancer}. For example, Cruz-Roa et al. used a CNN to examine primary breast cancer \cite{cruz2014automatic}, and Ertosun et al. studied the grading of gliomas \cite{ertosun2015automated}.

In this paper, we propose a framework for automatically locating cervical squamous tumor regions in WSIs and discerning tumor WSIs containing tumor regions from normal WSIs.  To the best of our knowledge, our work represents the first classifier to successfully applied on computer-aided cervical squamous cancer discrimination using WSIs. We used the most advanced CNN architectures and careful designed post-processing approach for the cervical tumor region location and slide-based classification. The results indicated that all the CNN models tested could identify tumors in WSIs well (the AUC scores $ > $0.94). And after ensemble model A to C, the performance on slide-based classification improved (AUC $ > $0.97 for model D).  

Although our platform was trained and validated properly, for the localization task of tumors in WSIs, our models only had moderate performance on FROC scores. It is a big challenge to train these models, because of the complexity and high Complexity and high variability of each WSI. Addition, the cervical cancer regions are much more complicated than other cancers, which characterized by small area, large number and scattered distribution. 

Recently, deep learning-based methods have consistently shown state of the art performance in pathology. Wang et al. Proposed an approach for identifying Metastatic breast cancer in 2016 \cite{wang2016deep}. Their slide level classification method attained  an AUC score of 0.925 and tumor region localization method achieved a FROC score of 0.7051. Dezso et al. introduced a method for detecting and classifying lesions in mammograms with deep learning with FROC score of 0.75 in 2018 \cite{ribli2018detecting}.The complexity of data which those literature used are different from ours. In most of their data, include training data and test data, tumor region is concentrated and results those data almost only have 1~3 tumor region. While our cervical cancer WSIs have $ > $300 tumor region. This makes our task of tumor region localization detection more difficult and also can further improve the efficiency of pathologists. 

In summary, we, for the first time, described a general deep learning platform for the diagnosis and referral of SCC. We focus on manually labeled histopathological WSIs that allow us to train an SCC-specific classifier. The approach we proposed can unfailingly recognize most tumor regions and the outputs were consistent across the same slides which scanned from various batches. We believe this fast and accurate method can be an invaluable tool for monitoring SCC evolution and greatly reduce the burden on pathologists

\appendix
\section{Acknowledgment}
This work is supported by the National Natural Science Foundation of China (No. 81872129).

\section{Appendix}
\setcounter{table}{0}
\renewcommand{\thetable}{S\arabic{table}}
\begin{table}
	\begin{center}
		\setlength{\tabcolsep}{1.2mm}{
			\begin{tabular}{lc}
				\toprule
				\textbf{Characteristics} & \textbf{Summary} \\ \hline
				Squamous Carcinoma of & \multirow{2}{*}{N=300} \\
				the Cervix Stage & \\
				\uppercase\expandafter{\romannumeral1} & N=2 (0.7\%) \\
				\uppercase\expandafter{\romannumeral1}A & N=16 (5.3\%) \\
				\uppercase\expandafter{\romannumeral1}A1 & N=4 (1.3\%) \\
				\uppercase\expandafter{\romannumeral1}A2 & N=0 (0.0\%) \\
				\uppercase\expandafter{\romannumeral1}B & N=23 (7.7\%) \\
				\uppercase\expandafter{\romannumeral1}B1 & N=108 (36.0\%) \\
				\uppercase\expandafter{\romannumeral1}B2 & N=19 (6.3\%) \\
				\uppercase\expandafter{\romannumeral2} & N=1 (0.3\%) \\
				\uppercase\expandafter{\romannumeral2}A & N=39 (13.0\%) \\
				\uppercase\expandafter{\romannumeral2}A1 & N=3 (1.0\%) \\
				\uppercase\expandafter{\romannumeral2}A2 & N=0 (0.0\%) \\
				\uppercase\expandafter{\romannumeral2}B & N=82 (23.3\%) \\
				\uppercase\expandafter{\romannumeral2}B1 & N=0 (0.0\%) \\
				\uppercase\expandafter{\romannumeral2}B2 & N=3 (1.0\%)\\
				\bottomrule&
		\end{tabular}}
	\end{center}
	\caption{Patient characteristics. }
	\label{tab:suplement}
\end{table}

{
\small
\bibliographystyle{ieee}
\bibliography{myref}
}

\end{document}